\documentclass{article}

\usepackage{amssymb}
\usepackage{amsmath}
\usepackage{latexsym}
\usepackage{algorithm}
\usepackage{algpseudocode}
\usepackage{multirow}
\usepackage{graphicx}
\graphicspath{{figures/}}
\usepackage{url}
\usepackage{xcolor}
\usepackage{subcaption}
\usepackage{mathtools}
\usepackage{booktabs}

\DeclareMathOperator*{\argmax}{argmax}

\begin{document}

\title{Drone swarm patrolling with uneven coverage requirements}
\author{C. Piciarelli, G.L. Foresti\\
University of Udine, Italy\\
claudio.piciarelli@uniud.it, gianluca.foresti@uniud.it}
\date{}


\maketitle

\begin{abstract}
Swarms of drones are being more and more used in many practical scenarios, such as surveillance, environmental monitoring, search and rescue in hardly-accessible areas, etc.. While a single drone can be guided by a human operator, the deployment of a swarm of multiple drones requires proper algorithms for automatic task-oriented control. In this paper, we focus on visual coverage optimization with drone-mounted camera sensors. In particular, we consider the specific case in which the coverage requirements are uneven, meaning that different parts of the environment have different coverage priorities. We model these coverage requirements with relevance maps and propose a deep reinforcement learning algorithm to guide the swarm. The paper first defines a proper learning model for a single drone, and then extends it to the case of multiple drones both with greedy and cooperative strategies. Experimental results show the performance of the proposed method, also compared with a standard patrolling algorithm.
\end{abstract}

\section{Introduction}
\label{sec:introduction}
During the last years, the use of multicopter drones has gained large popularity in many practical application fields, such as agriculture~\cite{tripicchio2015towards}, surveillance~\cite{motlagh2017uav}, disaster management~\cite{erdelj2017help}, search \& rescue~\cite{silvagni2017multipurpose}, environmental data acquisition~\cite{cruzan2016small}, medicine delivery~\cite{thiels2015use}, etc.. This interest in commercial, scientific, and social fields led to a drastic change in the way drones are used if compared to previous applications, which were mostly confined to video acquisition. While a single drone, manually controlled by a human operator, was a typical scenario up to few years ago, current applications often rely on multiple drones (swarms) autonomously cooperating to perform a given task. This justifies the increment of scientific works on topics such as automatic drone control, path planning, smart resource management, etc..

In this work, we focus on the problem of drone visual coverage: multicopter drones are equipped with cameras to observe a portion of the environment for specific tasks (e.g. surveillance) and the observed area must be optimized according to given criteria, under the assumption that the environment is too large to get a simultaneous full coverage. For example, in a surveillance application, a basic requirement could be that no area is left uncovered for more than a given amount of time, in order to avoid ``blind spots'' in the surveillance application.

A task like the one just described can be accomplished with uniform coverage, given that enough drones are available. In this paper, however, we want to investigate scenarios with uneven coverage requirements. This means that not all the areas of the environment have the same importance: some parts should be observed more frequently, or require immediate coverage, if compared to other zones. Each point of the environment thus has a given priority (or \emph{relevance}, using the term that will be adopted in the rest of the paper), and the goal of the system is to find a patrolling strategy that optimally covers the environment according to these relevance requirements. For example, in a disaster management context such as a flooding or fire in extended rural areas, the zones around known buildings should have priority for visual inspection in order to quickly identify people in danger.

We propose to model the problem of prioritized visual coverage as a Markov Decision Process, where each drone is an agent that can actuate several actions to change its state (e.g. moving forward, rotating, zooming the camera, etc.) and it gets a reward (either positive or negative) in terms of coverage quality as a consequence of its actions. This model allows to search for a patrolling strategy using reinforcement learning, thus letting the drone learn from experience rather than explicitly giving a patrolling algorithm. The reinforcement learning algorithm will be implemented using a neural network, thanks to the Deep Q-Network architecture. The proposed model is theoretical, meaning that the set of actions and the state space do not consider all the variables of a real-world system (such as drone movement inertia, power consumption constraints, etc.) however it can be used as a reference framework for possible implementations. Our basic idea of reinforcement-learning-based patrolling was already presented in~\cite{piciarelli19-drone}, however in that work only the single-drone case was considered. In this paper we extend our previous work by improving the patrolling model (see eq.~\ref{eq:singletemporalmask}) and by considering the novel scenario of multi-drone swarms. In this case we propose two different strategies, a greedy and a cooperative one, depending on the type of information shared between drones. Compared to~\cite{piciarelli19-drone}, also the experimental results have been extended, even for the single-drone scenario.


The paper is organized as follows: in Section~\ref{sec:sota} we review some of the most relevant works in literature regarding drone coverage tasks, while in Section~\ref{sec:rl} we recap the basic theory behind reinforcement learning and its deep learning implementations. In Section~\ref{sec:singledronemodel}, the Markov Decision Process model for a single drone is described by defining the state space, the action space, and the reward function in terms of visual coverage of relevance maps. A patrolling strategy is then implemented using the given model, which is then extended to the case of multiple drones in Section~\ref{sec:swarm}, where two swarm patrolling strategies are proposed. Experimental results are given in Section~\ref{sec:results}.

\section{Related works}
\label{sec:sota}

The problem of drone control for patrolling tasks has been studied by several authors. A survey of coverage path planning algorithms in robotics can be found in~\cite{galceran2013survey}, while~\cite{cabreira2019survey} is specifically focused on drones. Here, the authors propose a taxonomy of the cited methods, ranging from simple geometric flight patterns to more complex grid-based solutions considering full and partial information about the area of interest. The authors mostly focus on how to cover areas with complex geometrical shapes but, in contrast with the proposed method, none of the surveyed works explicitly considers uneven coverage requirements, neither addresses the problem using neural networks.

In~\cite{Piciarelli2016-dynamic}, the authors give a survey on dynamic reconfiguration of camera networks, which is a superset of the considered problem. They explicitly discuss coverage optimization methods and drone deployment strategies, although the two topics are separated: surveyed coverage-oriented methods are mostly focused on PTZ camera networks (in which the camera cannot translate), while the analyzed drone reconfiguration works are more focused on resource management (e.g. as in~\cite{yanmaz2018drone,wischounig-Strucl2015resource}). 

The joint task of area coverage and resource management has been studied in~\cite{bentz20173d}, where the authors give a deep mathematical formulation of 3D coverage and they propose a resource-aware algorithm that shift the bulk of spatial redistribution onto less constrained agents. A similar topic is addressed in~\cite{difranco2015energy}, where the problem of coverage-driven path planning is studied from the point of view of resource management such as energy consumption. In~\cite{Curiac2015path}, the problem is analyzed from the novel point of view of path planning in adversarial environments, where the efficient use of chaotic behaviors copes with enemy entities. 
The work proposed in~\cite{liu2018energy} is again oriented to energy-efficient algorithms for coverage optimization, although in this case the authors deal with communication coverage, where drones are used as a communication infrastructure to deal with emergency situations.
Other works focus on the decentralized aspects of the task, such as in~\cite{zargar2016decentralized}: in the case of a swarm of drones, distributing the overall computation over all the agents allows efficient implementations that do not rely on a single point of failure. All these works are thus mostly focused on resource awareness, a topic that we do not address in this work.

To the best of our knowledge, few works have been published dealing with drone coverage problems using neural networks and/or reinforcement learning. The work presented in~\cite{luo2018deep} uses a deep-sarsa approach, thus adopting a reinforcement-learning-based approach as in our work, however it is focused on target-based guidance with collision avoidance rather than on patrolling. In~\cite{koch2019reinforcement}, the authors use reinforcement learning for attitude control, and thus they are more focused on short-term stability-oriented tasks rather than mission-level, long-term objectives like in our case.

\section{Reinforcement learning essentials}
\label{sec:rl}
Reinforcement learning is a learning strategy in which an agent in a given state executes an action and gets an immediate reward (or penalty) as a consequence of that action. The goal is to learn from this experience, figuring out the best actions that will eventually lead to a maximization of the total reward on the long term. This approach can be easily applied to coverage-oriented drone patrolling problems, since the drone (the agent) can take several actions (e.g. moving forward, zooming in, etc.) that will impact its visual coverage of the environment. It must be noted that trivial solutions, such as choosing the action that will immediately maximize the coverage, are not suited for problems with uneven coverage requirements. In this case, the best short-term action does not necessarily lead to a good long-term solution, e.g. when the drone must necessarily  cross a low-priority area in order to reach an high-priority zone. This justifies the use of reinforcement learning as a technique to find the optimal patrolling strategy.

Formally, reinforcement learning models the problem as a Markov Decision Process $\mathcal{P} = \{S, A, \tau, r, \gamma \}$, where $S$ is a finite set of agent states and $A$ is a finite set of actions. $\tau(s'|s,a)$ is the transition probability from state $s$ to state $s'$ given that action $a$ is executed, $r(s,a)$ is the reward obtained by executing action $a$ in state $s$ and $\gamma$ is a discount factor, modeling the importance of immediate, short term rewards with respect to past rewards. A fundamental property of Markov Decision Processes is that the transition probability is defined only in terms of $s,s'$ and $a$, thus meaning that the next state will be affected only by the current state and action, and not by the history of previous states. A \emph{policy} is a probability function $\pi(a|s)$ denoting the probability for an agent in state $s$ to take action $a$, and the goal of reinforcement learning is to find the optimal policy $\pi^*$ that maximizes the expected total discounted reward:
\begin{equation}
\pi^* = \argmax_\pi  \mathbb{E}_{\tau, \pi} \left[ \sum_{i=0}^\infty \gamma^i r(s_i, a_i) \right]
\label{eq:optimalpolicy}
\end{equation}
where $\mathbb{E}_{\tau, \pi}$ means that the expected value is computed assuming that the states sequence is distributed according to $\tau$ and the actions are chosen according to $\pi$. Several methods have been proposed to solve (\ref{eq:optimalpolicy}), one of the most popular being \emph{Q-learning}.

\subsection{Q-learning}
\label{subsec:qlearning}
The Q-learning algorithm is based on the definition of a function $Q_\pi$ modeling the total discounted reward that can be achieved following the policy $\pi$ if action $a$ is chosen in state $s$:
\begin{equation}
Q_\pi (s,a) = \mathbb{E}_{\tau,\pi,s_0=s,a_0=a} \left[ \sum_{i=0}^\infty \gamma^i r(s_i,a_i) \right].
\label{eq:qvalue}
\end{equation}
Equation (\ref{eq:qvalue}) can be written recursively as:
\begin{equation}
Q_\pi (s,a) = r(s,a) + \gamma \sum_{s',a'} \tau(s'|s,a) \pi(a',s') Q_\pi(s',a')
\label{eq:qrecursive}
\end{equation}
and it can be simplified if the optimal policy $\pi^*$ is considered. In fact, $\pi^*(a,s)$, because of its optimality, has a binary nature: it has value 1 for the best action possible and 0 for any other action, thus the $Q^*$ function for the optimal policy reduces to:
\begin{equation}
Q^*(s,a) = r(s,a) + \gamma \sum_{s'} \tau(s'|s,a) \max_{a'} Q^*(s',a')
\label{eq:qstar}
\end{equation}
If we restrict our analysis to deterministic problems, in which also $\tau(s'|s,a)$ can assume only binary values, eq. (\ref{eq:qstar}) further simplifies to the so-called Bellman equation~\cite{russell2016artificial}:
\begin{equation}
Q^*(s,a) = r(s,a) + \gamma \max_{a'} Q^*(s', a')
\label{eq:bellman}
\end{equation}
where $s'$ is the state reached from $s$ by executing action $a$, in other words it is the only state such that $\tau(s'|s,a)=1$. 

The Q-learning algorithm computes $Q^*$ using eq. (\ref{eq:bellman}) and dynamic programming. It starts by filling a $|S|\times |A|$ table with initial random values $Q^*_0(s,a)$ for each possible combination of states and actions, and then updates them iteratively until convergence according to the following equation:
\begin{equation}
Q^*_{i+1}(s,a) = r(s,a) + \gamma \max_{a'} Q^*_i(s', a') .
\label{eq:finalq-iter}
\end{equation}
In order to apply (\ref{eq:finalq-iter}), Q-learning requires the knowledge of the reward $r(s,a)$. This knowledge comes from experience: in the training phase, the agent is requested to perform actions in order to measure the corresponding reward and accumulate enough data to update $Q^*$ until convergence. In theory, the actions could be always chosen randomly, but in practice it is more effective to draw some of them from the partially-learned policy in order to evaluate the most promising parts of the state-action space. This approach is called exploration-exploitation strategy, where exploration refers to the random choice of actions to explore previously unseen state-action pairs and exploitation refers to exploiting the current estimate $Q^*_i$ in order to choose the action that maximizes it in the current state. A typical training starts with exploration only, and the exploitation becomes more and more frequent as the iterative process progresses.

Once $Q^*$ is computed, finding the optimal policy is trivial, since the best action to be taken in state $s$ is the one that maximizes $Q^*(s,\cdot)$.

\subsection{Deep Q-Networks}
\label{subsec:deepqnet}
The dynamic programming approach described in section~\ref{subsec:qlearning} requires the memorization of a $|S|\times |A|$ table which is often impractical, since |S| could be large (or even infinite, if we extend the definition of Markov Decision Processes to infinite state spaces, e.g. with continuous rather than discrete values). Deep Q-Networks approaches~\cite{mnih2015human,mnih-atari-2013} solve the problem by using deep neural networks as function approximators for $Q^*$. 

Let $Q^*(s;\theta)$ be a neural network with parameters $\theta$, taking as input a state $s$ and giving as output the values $Q^*(s;\theta)|_k, k\in\{1\ldots |A|\}$ representing the $Q^*$ value for each possible action in state $s$. Then, when an experience tuple $(s,a,r,s')$ is acquired, it can be used to train the network by minimizing the following MSE loss function:
\begin{equation}
L(\theta) = \mathbb{E}\left[ \left(Y - Q^*(s;\theta)|_a\right)^2 \right]
\label{eq:lossfunction}
\end{equation}
representing the difference between the current estimate $Q^*(s;\theta)|_a$ of the value of action $a$ in state $s$ and the new estimate that can be computed from the experience, defined as:
\begin{equation}
	Y = r(s,a) + \gamma \max_{k}Q^*(s';\theta)|_k.
	\label{eq:Y}
\end{equation}

The computation of the loss function~\ref{eq:lossfunction} can lead to bias issues in practical implementations. Neural networks are typically trained in small batches, and if the experience data of a batch come from the same experiment (e.g. subsequent steps of the same agent) they can lead to biased computation of the expected value. In order to avoid this problem, a \emph{replay memory} can be used, which consists in a large set of experience tuples. During the training phase the tuples, rather than being directly used to train the network, are stored in the replay memory. Batches are then built by sampling the replay memory with uniform distribution, thus avoiding to build batches composed only of highly  correlated tuples.

Another problem in the computation of the loss function comes from the network parameters $\theta$ which are used both for action selection in eq.~(\ref{eq:Y}) and for action evaluation in eq.~(\ref{eq:lossfunction}). It has been proven that this could lead to biased results~\cite{van2016deep}, which can be avoided by decoupling selection and evaluation using two different networks. Two of the most popular approaches are the Target Network approach~\cite{mnih2015human}, in which the new estimate is defined as:
\begin{equation}
Y= r(s,a) + \gamma \max_{k} Q^*(s';\theta^-)|_k
\label{eq:targetnetwork}
\end{equation}
and the Double DQN approach~\cite{van2016deep}:
\begin{equation}
Y = r(s,a) + \gamma Q(s';\theta^-)|_{\argmax_k Q(s'; \theta)|_k}
\label{eq:doubledqn}
\end{equation}
where $\theta^-$ are the parameters of a second deep Q-network. This second network, rather than being trained independently, is generally defined in terms of the first network, either via hard update ($\theta^-$ is set to $\theta$ every a fixed number of epochs) or via soft update at each epoch $i$, according to a temporal smoothness factor $\alpha\in[0,1]$:
\begin{equation}
	\theta^-_i = (1-\alpha)\theta^-_{i-1} + \alpha \theta_i.
	\label{eq:softupdate}
\end{equation}

\section{Single drone model}
\label{sec:singledronemodel}
In order to apply reinforcement learning techniques to drone patrolling tasks, we must model the drone as a Markov Decision Process agent. This implies defining its possible states, the actions and a reward function giving a positive or negative feedback to each action.

\subsection{State space}
\label{subsec:statespace}
The state space should consider all the relevant parameters that define the drone setup at a given time instant. We identified six parameters that directly influence the visual coverage of a drone. Formally, a drone state is a tuple $s=\{x,y,z,\psi,\phi,f\}$ defined as:
\begin{itemize}
	\item $x, y, z$: spatial coordinates of the drone;
	\item $\psi$: camera orientation angle;
	\item $\phi$: camera tilt angle;
	\item $f$: camera focal length.
\end{itemize}

The spatial coordinates $x,y,z$ are referred to a world reference system, and are limited by the borders of the area to be monitored and by the maximum flying height the drone can reach. The camera orientation angle $\psi\in[0,2\pi]$ is the camera azimuth, expressed as the angle between the camera frontal axis and the $X$ axis of the world reference system. The camera tilt angle $\phi\in [0,\pi/2]$ describes the elevation of the camera, where $\phi=0$ is the camera pointed at the horizon and $\phi=\pi/2$ is a nadiral view. Finally, the focal length $f$ is included in the state  to model zoom cameras, and its range is hardware dependent. 

Observe that we did not include the drone azimuth in the state space, as this information is not relevant. The proposed work is focused on multicopter drones, which can move freely in the three spatial dimensions (as opposed to fixed-wing drones), thus identifying a frontal axis is unnecessary: in the proposed framework, a drone aiming north and moving frontward is equivalent to a drone aiming east and moving leftward.

\subsection{Action space}
\label{subsec:actionspace}
As done for the state space, we identified a a set of drone actions that directly influence the visual coverage of the drone. The action space consists of a total of 12 actions:
\begin{itemize}
	\item Move \{Forward | Backward | Left | Right | Up | Down \};
	\item Rotate \{Left | Right\};
	\item Tilt \{Down | Up\};
	\item Zoom \{In | Out\}.
\end{itemize}

The \emph{Move} actions translate the drone in the 3D space and influence the $\{x,y,z\}$ components of the drone state. Without loss of generality, the front direction is assumed to be the camera orientation angle: as described in~\ref{subsec:statespace}, there is no need to decouple drone and camera azimuths. The camera orientation is defined by the \emph{Rotate} and \emph{Tilt} actions, which respectively influence the $\{\psi, \phi\}$ parameters. Finally the \emph{Zoom} actions change the focal length $f$ and thus the zoom level of the camera.

Note that the proposed actions do not quantify the amount of requested change, e.g. how much the drone should move when a MoveForward action is executed. Ideally those values should be continuous, and the action space would be infinite. However, the reinforcement learning techniques described in section~\ref{sec:rl} require a finite action space, and thus the actions must be discretized. The amount of parameter change caused by each action is thus fixed and defined a priori. For example, the experimental results discussed in section~\ref{sec:results} have been obtained with a \emph{Rotate} step of $\pi/16$, meaning that the camera performs a full $2\pi$ rotation after 32 \emph{Rotate} actions in the same direction.

\subsection{Visual coverage}
\label{subsec:coverage}
In order to evaluate the visual coverage quality of a drone, we need a way to compute the portion of the environment observed by the camera.
\begin{figure}
	\centering
	\includegraphics[width=0.8\linewidth]{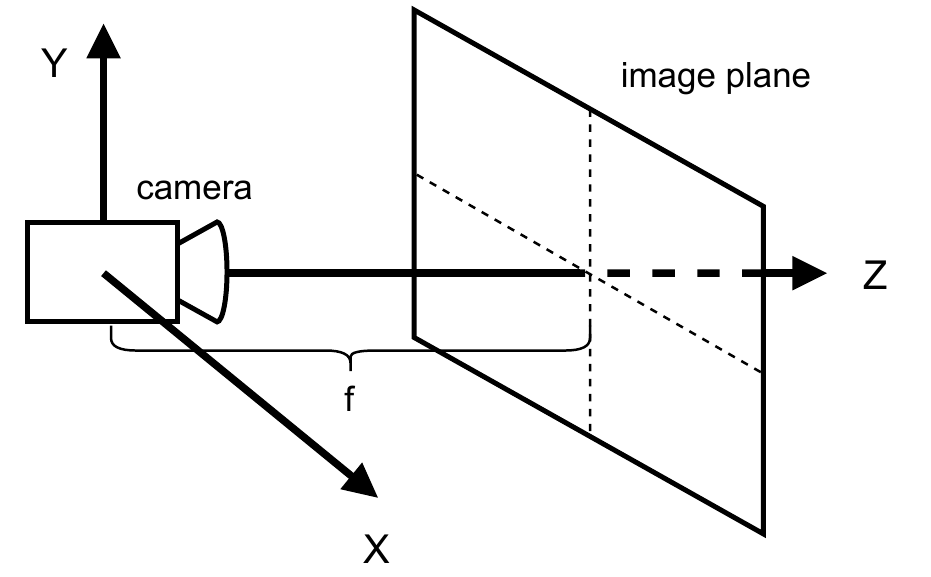}
	\caption{Camera reference system.}
	\label{fig:imageplane}
\end{figure}
Let us consider a reference system as the one depicted in Figure~\ref{fig:imageplane}, where the origin lies on the center of the camera optics, the $Y$ axis points upwards and the $Z$ axis is initially aligned with the camera optical axis when the camera points at the horizon with no rotation. The effect of actions \emph{Rotate \{Left | Right\}} can be modeled by a rotation matrix $R_\psi$ around the $Y$ axis: 
\begin{equation}
	R_\psi = \begin{bmatrix}
	\cos \psi & 0 & \sin \psi \\
	0 & 1 & 0 \\
	-\sin \psi & 0 & \cos\psi
	\end{bmatrix}
\end{equation}
while the actions \emph{Tilt \{ Up | Down\}} are modeled by a rotation around the $X$ axis:
\begin{equation}
	R_\phi = \begin{bmatrix}
	1 & 0 & 0 \\
	0 & \cos \phi & \sin \phi \\
	0 & \sin \phi & \cos \phi
	\end{bmatrix}.
\end{equation}

Any camera configuration can thus be modeled by a combination $R_{\psi\phi}$ of the two matrices, observing that the tilt rotation must be applied first, in order to correctly model the motion of a gimbal camera: 
\begin{equation}
	R_{\psi\phi} = R_\psi R_\phi = \begin{bmatrix}
	\cos\psi & \sin\psi \sin\phi & \sin\psi \cos\phi \\
	0 & \cos\phi & \sin\phi \\
	-\sin\psi & \sin\phi\cos\psi & \cos\phi \cos\psi
	\end{bmatrix}
	\label{eq:rotpsiphi}.
\end{equation}

By using the matrix~(\ref{eq:rotpsiphi}), it is possible to project on the ground plane any point on the image plane, given the camera rotation and tilt angle. Let $p = (x,y)$ be a pixel in the image plane. According to the camera pinhole model, the focal length $f$ is the distance between the optics and the image plane itself, and thus the coordinates of the pixel in the camera reference frame are $\hat{p}=(\delta_x x, \delta_y y, f)$, where $\delta_x, \delta_y$ are the pixel sizes of the imaging sensor. If the camera is rotated and tilted, point $\hat{p}$ is moved to $\hat{p}_{\psi\phi} = R_{\psi\phi}\hat{p}$. By switching to a world reference system, where the camera has coordinates $C = (C_x, C_y, C_z)$, the parametric equation of a line parallel to vector $\hat{p}_{\psi\phi}$ and passing through $C$ is:
\begin{equation}
	L(t) = C + t \hat{p}_{\psi\phi}.
	\label{eq:line}
\end{equation}

To find the intersection of $L$ with the ground plane it is sufficient to set the $Y$ world coordinates to zero:
\begin{equation}
	\begin{gathered}
	C_y + t \hat{p}_{\psi\phi, y} = 0 \\
	t = -C_y / \hat{p}_{\psi\phi, y}
	\end{gathered}
	\label{eq:t_togroundplane}
\end{equation}
and by substituting (\ref{eq:t_togroundplane}) in (\ref{eq:line}) we get the projection $p_g$ of point $p$ on the ground plane:
\begin{equation}
	p_g = C - \frac{C_y}{\hat{p}_{\psi\phi,y}} \hat{p}_{\psi\phi}.
	\label{eq:projfinal}
\end{equation}

By using~(\ref{eq:projfinal}), it is possible to project on the ground plane any point $p$ in the image plane given the camera angles $\psi, \phi$. If the equation is applied to the four corners of the image plane, it is thus possible to compute the corners of the trapezoidal projection of the image plane on the ground plane, which is the portion of the environment observed by the camera (we here assume that no point lies above the horizon). We define this zone the \emph{visual coverage} of the drone in state $s$, for now on denoted as $V(s)$.

\subsection{Reward function}
\label{subsec:reward}
As stated in section~\ref{sec:introduction}, this works deals with the case of uneven coverage requirements. This means that not all the portions of the environment have the same priority, and some areas are more important than others and require immediate visual inspection. In~\cite{piciarelli2011-automatic}, the authors used this approach to focus a surveillance system on the areas with highest activity, while in~\cite{piciarelli2012-network} the high priority areas were identified by audio sensors. In general, the definition of these areas is extremely context-dependent, and here we just assume that such a definition exists in the form of a \emph{relevance map}. A relevance map $\mathcal{M}(x,y): \mathbb{R}^2 \to [0,1]$ is thus a function taking as input the $(x,y)$ coordinates of a point in the world reference system and returning a value in the range $[0,1]$ denoting the relevance of that point, i.e. the relative importance of getting visual coverage of that point with respect to the rest of the map.

It is now possible to define the \emph{observed relevance} $\rho(s)$ of a drone in state $s$ as the total relevance within its visual coverage:
\begin{equation}
	\rho(s) = \int\!\int_{V(s)} \mathcal{M}(x,y)\ dA
\end{equation}
or, in the likely case that $\mathcal{M}$ is discretized in a matrix:
\begin{equation}
	\rho(s) = \sum_{\mathclap{(x,y)\in V(s)}} \mathcal{M}(x,y)
	\label{eq:rho}
\end{equation}
where $V(s)$ is the visual coverage of the drone, as defined in section~\ref{subsec:coverage}.

\begin{figure}
	\centering
	\includegraphics[width=0.7\linewidth]{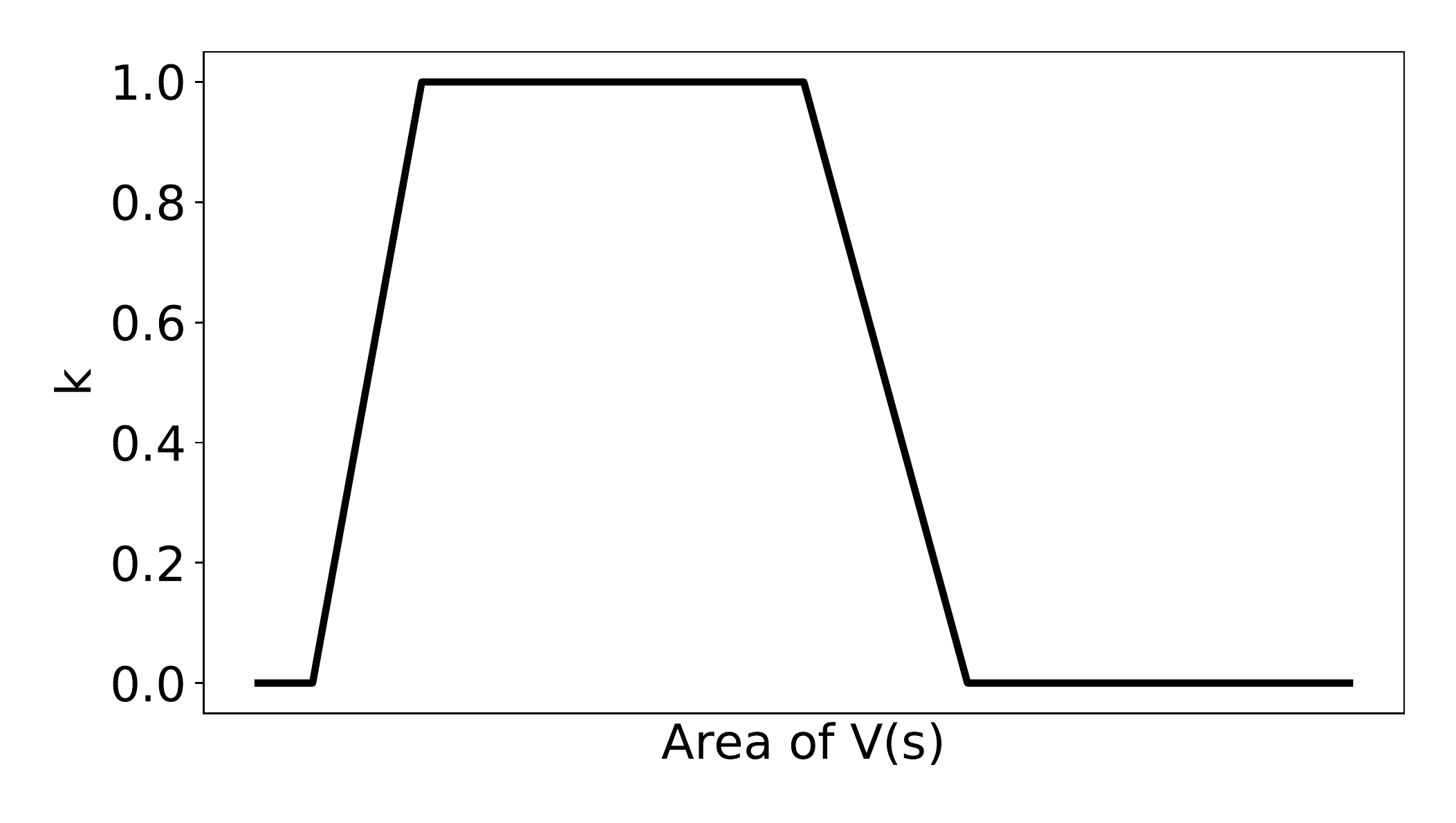}
	\caption{Penalty function to penalize large visual coverage.}
	\label{fig:penalty}
\end{figure}
The drone reward function could thus be defined in terms of its observed relevance, giving a positive reward to actions that increase $\rho(s)$. However, additional constraints are needed in order to avoid extreme cases, such as the drone flying at the maximum possible height trying to cover the entire area. Despite in small areas this could be a viable solution, in larger environments it is most probably a useless configuration from a practical point of view because of the very low spatial resolution (pixel per meter) of the acquired images. We thus enforce a constraint on the size of the visual coverage by defining a penalty function $k:\mathbb{R}\to [0,1]$ such as the one shown in figure~\ref{fig:penalty} that penalizes coverages which are either too small or too large for practical applications; the shape and tuning of $k$ is of course application dependent. We thus define the \emph{Constrained Observed Relevance} (COR) as:
\begin{equation}
	\hat{\rho}(s) = k\left(V(s)\right) \rho(s).
	\label{eq:cor}
\end{equation}

The drone reward function $r(s,a)$ can now be defined as:
\begin{equation}
	r(s,a) = \hat{\rho}(s') - \hat{\rho}(s)
	\label{eq:correward}
\end{equation}
this is, if the execution action $a$ in state $s$ leads to state $s'$, the corresponding reward is defined as the difference of Constrained Observed Relevance between the two states $s'$ and $s$. By using such a function, the reinforcement learning system is positively rewarded each time the agent chooses actions that lead to an increase of observed relevance. Since reinforcement learning algorithms maximize the total reward on the long term, this means that the drone will try to cover high-relevance areas, even if this requires to move through low-relevance zones in the short term.

\subsection{Patrolling}
\label{subsec:patrolling}
Training a reinforcement learning network with the reward function described in~\ref{subsec:reward} is not sufficient to get a patrolling algorithm. The reward function just forces the drone to move to more relevant areas in order to increase the total reward, and the process will stop once the highest possible value has been reached: the algorithm is actually a state-space explorer trying to find a path to the global maximum of the COR function.

In order to get a sensible patrolling behavior, the temporal aspect must be introduced. More specifically, we assume that an area that is under observation by the drone should have its relevance reduced, in order to give the drone the opportunity to move to other still unexplored zones. At the same time, it is reasonable to require that, given two areas with the same relevance, higher priority should be given to the one that has been unobserved for the longest time. 

In order to model the temporal aspect, we introduce the \emph{temporal relevance map} 
\begin{equation}
\mathcal{M}^t_s (x,y) = \mathcal{M}(x,y) T^t_s(x,y)
\label{eq:temporalrmap}
\end{equation}
defined as a combination of the static map $\mathcal{M}$ and a temporal mask $T^t_s: \mathbb{R}^2\to [0,1]$. The temporal mask at time instant $t$ for a drone in state $s$ is defined as:
\begin{multline}
T^t_s(x,y) = \\
\begin{cases}
	1 & \text{for } t=0 \\
	\min(1, T^{t-1}_s(x,y) + \delta_+) & \text{for } t>0 \land (x,y) \not\in V(s) \\
	\max(0, T^{t-1}_s(x,y) - k(V(s))\delta_-) & \text{for } t>0 \land (x,y) \in V(s) 
\end{cases}
\label{eq:singletemporalmask}
\end{multline} 
where $k$ is the penalty function defined in section~\ref{subsec:reward}, $V(s)$ is the visual coverage of the drone in state $s$ defined in section~\ref{subsec:coverage} and $\delta_-, \delta_+$ are constant decreasing and increasing factors. The decreasing factor is multiplied by the penalty function $k(V(s))$ because we do not want to model the contribution of drones that do not satisfy the area constraints, as already done in the definition of the Constrained Observed Relevance (eq.~\ref{eq:cor}).

With this definition, the temporal relevance map $\mathcal{M}^t_s$ dynamically changes depending on the time since last observation for each point of the map. Applying the reinforcement learning algorithm on this map forces the drone to continuously move around the area searching for high-relevance areas that have not been observed since a long time, thus implementing an efficient patrolling algorithm.

Finally, observe that the temporal map should now be part of the state, since it is dependent on the chosen actions rather than being static. The reward function~(\ref{eq:correward}) should now be rewritten as 
\begin{equation}
	r(\{s,\mathcal{M}_s^t\},a) = \hat{\rho}(\{s',\mathcal{M}_{s'}^{t+1}\})-\hat{\rho}(\{s,\mathcal{M}_s^{t}\})
	\label{eq:newreward}
\end{equation}
and the definitions of $\hat{\rho}, \rho$ are consequently adapted so that the summation in eq.~(\ref{eq:rho}) is performed over $\mathcal{M}_s^t$ rather than $\mathcal{M}$.

\section{Swarm model}
\label{sec:swarm}
The model proposed in section~\ref{sec:singledronemodel} describes how a single drone can be modeled as a Markov Decision Process agent, and its patrolling strategy can be defined using reinforcement learning and a proper reward function. However, the model can be extended to a swarm of drones, where multiple drones share the task of patrolling a given area trying to maximize the overall visual coverage of relevant areas. We here propose two approaches, the \emph{greedy} strategy and the \emph{cooperative} one.

\subsection{Greedy strategy}
\label{subsec:greedy}
The greedy strategy consists in applying the single-drone model to each drone of the swarm. This way, every agent will try to maximize its own total reward in a greedy way, this is independently from other swarm members. However, a naive application of this strategy would eventually lead all the drones to cover the same areas of the map, namely the ones with highest relevance. In order to turn the greedy strategy into a sensible patrolling algorithm, it is sufficient to request that the temporal relevance map is shared among all the swarm. With map sharing, each drone will naturally avoid the areas already observed by other members of the swarm because the temporal relevance of those areas will be decreasing due to coverage. The approach is still greedy since each drone does not explicitly know about the presence of other drones in the area, however this knowledge is indirectly modeled by the shared map. 

The shared temporal relevance map for a swarm of $N$ drones with states $\mathbf{S}=\{s_i\}_{i=1}^N$ is defined as:
\begin{equation}
	\mathcal{M}^t_\mathbf{S}(x,y) = \mathcal{M}(x,y) T^t_\mathbf{S}(x,y)
\end{equation}
where the temporal mask follows the same principle of equation~(\ref{eq:singletemporalmask}), but extended to all the drones in the swarm:
\begin{multline}
T^t_\mathbf{S}(x,y) = \\
\begin{cases}
	1 & \text{for } t=0 \\
	\min\big(1, T^{t-1}_\mathbf{S}(x,y) + \delta_+\big) & \text{for } t>0 \land (x,y) \not\in \bigcup\limits_{s\in\mathbf{S}} V(s) \\
	\!\begin{split}
		\max\Big(0, T^{t-1}_\mathbf{S}(x,y) - &\\
		\sum_{\mathclap{s|(x,y)\in V(s)}} k(V(s))\delta_- \Big) &
	\end{split}
	& \text{for } t>0 \land (x,y) \in \bigcup\limits_{s\in\mathbf{S}} V(s) 
\end{cases}.
\label{eq:multitemporalmask}
\end{multline} 
This way, the temporal mask increases if the point $(x,y)$ is not observed by any drone, but it is decreased multiple times if it is observed by several drones, to encourage the multiple coverage of zones where the relevance is particularly high. The reward function~(\ref{eq:newreward}) is consequently redefined as:
\begin{equation}
r(\{s,\mathcal{M}_\mathbf{S}^t\},a) = \hat{\rho}(\{s',\mathcal{M}_{\mathbf{S}'}^{t+1}\})-\hat{\rho}(\{s,\mathcal{M}_\mathbf{S}^{t}\})
\label{eq:greedyreward}
\end{equation}

The computation of the shared map can be done by a central processing node and sent at each time interval to all the swarm. This is particularly suitable if the entire computation is done offline: in this case the algorithm is just used to pre-compute a patrolling strategy which is subsequently executed. If the algorithm must be applied online, a distributed approach would be preferred; this case is covered in the next section.

\subsection{Cooperative strategy}
\label{subsec:coop}
In the case of cooperative strategy, each drone is aware of the rest of the swarm and their states, so that collaborative strategies can be implemented directly, rather than indirectly via the shared map as in the greedy strategy. However, rather than explicitly define the cooperative models, we rely on reinforcement learning to automatically learn them from experience. 

\begin{figure}
	\centering
	\includegraphics[width=0.9\linewidth]{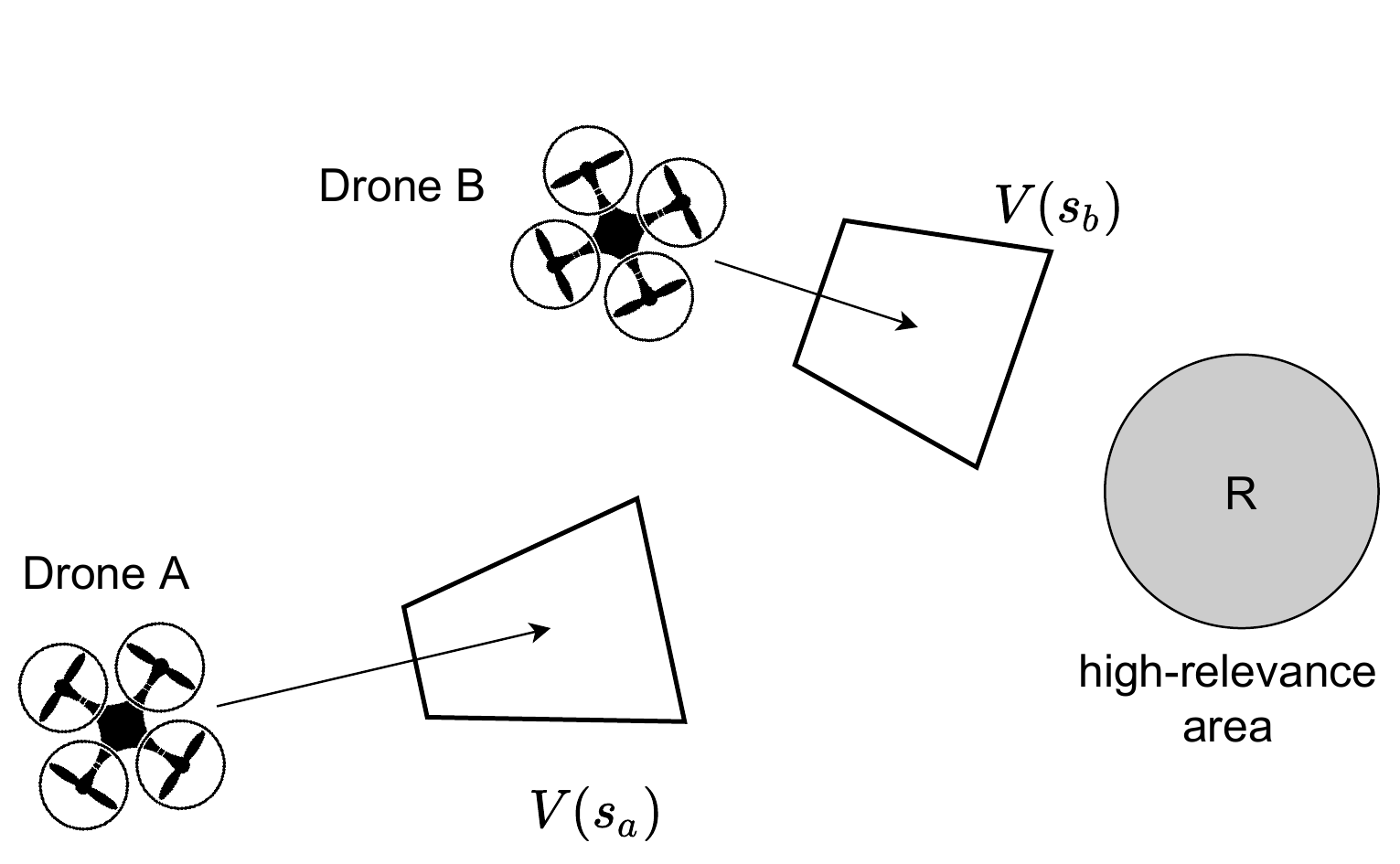}
	\caption{Two competing drones for the same high-relevance area.}
	\label{fig:2drones}
\end{figure}

The idea is to use $\mathbf{S} = \{s_i\}_{i=1}^N$, the set of all the states of the drones, as a new global state. The implementation is straightforward, since the reward function for agent in state $s$ and taking action $a$ leading to state $s'$ is defined as:
\begin{equation}
	r(\{\mathbf{S}, \mathcal{M}_\mathbf{S}^t\}, a) = 
	\hat{\rho}(\{s',\mathcal{M}_{\mathbf{S}'}^{t+1}\})-\hat{\rho}(\{s,\mathcal{M}_\mathbf{S}^{t}\})
	\label{eq:coopreward}
\end{equation}
However, this is the same as equation~(\ref{eq:greedyreward}), except for $\mathbf{S}$ being made explicit in the function input. The difference between the cooperative and greedy strategy in fact does not lie in the reward function, but in the way rewards are related to states, and this is what reinforcement learning tries to learn. The Q-function (see section~\ref{subsec:qlearning}) learned by the greedy approach is $Q(\{s,\mathcal{M}_\mathbf{S}^t\}, a)$ while the one learned by the cooperative approach is $Q(\{\mathbf{S},\mathcal{M}_\mathbf{S}^t\}, a)$. In order to clarify the difference, consider the example of Figure~\ref{fig:2drones}, showing two drones $A$ and $B$, their visual coverages $V(s_a), V(s_b)$ and an high-relevance area $R$. With the greedy strategy, both drones will move toward $R$, however $B$ will reach it first since it is closer. Once $B$ covers $R$, the shared temporal relevance will decrease because of temporal update, and $A$ will probably switch to another, more relevant target. With the cooperative approach instead, by explicitly knowing the state of $B$, drone $A$ could predict that $R$ will be covered by the closer drone and immediately switch to another target. In other words, despite the reward function not changing between the two algorithms, the greedy approach relies on \emph{observing} changes in the temporal map $\mathcal{M}^t_\mathbf{S}$, while the cooperative approach could \emph{predict} them thanks to the explicit knowledge of $\mathbf{S}$. Of course the example shows that this form of reasoning is possible, but since we do not model any explicit cooperative strategy, we need to prove with experimental results that reinforcement learning can automatically infer this kind of strategies from experience.

Regarding practical implementations, each drone needs to know the state of the entire swarm at each time interval. This could be achieved by using a central processing node, as already discussed in section~\ref{subsec:greedy}, and the knowledge of $\mathbf{S}$ can also be used by each drone to independently reconstruct the entire map $\mathcal{M}_\mathbf{S}^t$. A distributed approach could be implemented by drones exchanging state vectors each other when they are within communication range, thus propagating the state information across the entire swarm network, however this solution is not detailed here.

\subsection{Deep learning implementation}
\label{subsec:deepimplementation}
In order to implement the greedy and cooperative techniques, we adopted the Deep Q-Network approach described in section~\ref{subsec:deepqnet} with a linearly decreasing ratio between exploration and exploitation steps.  In order to avoid training biases, the Double DQN model has been used (eq.~\ref{eq:doubledqn}) with replay memory and soft update of the second network. 

\begin{figure*}[tbh]
	\centering
	\begin{subfigure}[t]{0.25\textwidth}
		\centering
		\includegraphics[width=0.9\textwidth]{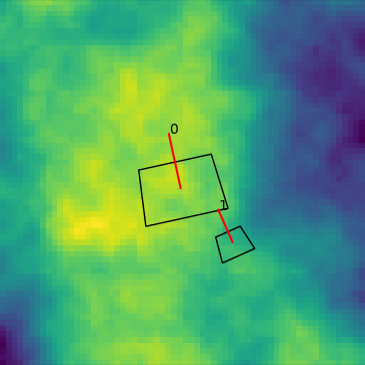}
		\caption{\label{fig:input:1}}
	\end{subfigure}%
	\begin{subfigure}[t]{0.25\textwidth}
		\centering
		\includegraphics[width=0.9\textwidth]{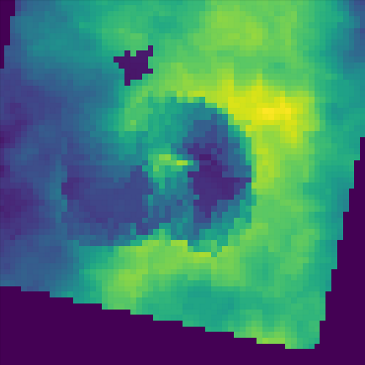}
		\caption{\label{fig:input:2}}
	\end{subfigure}%
	\begin{subfigure}[t]{0.25\textwidth}
		\centering
		\includegraphics[width=0.9\textwidth]{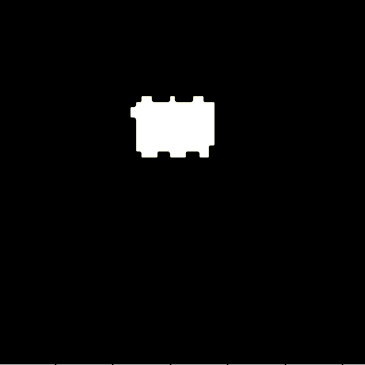}
		\caption{\label{fig:input:3}}
	\end{subfigure}%
	\begin{subfigure}[t]{0.25\textwidth}
		\centering
		\includegraphics[width=0.9\textwidth]{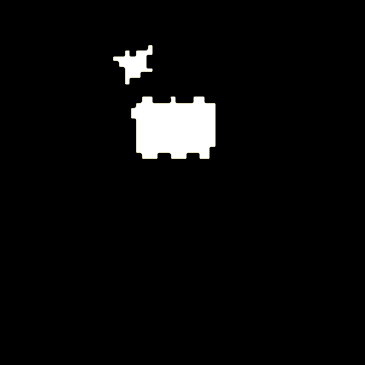}
		\caption{\label{fig:input:4}}
	\end{subfigure}%
	\caption{Network input example. (\subref{fig:input:1}) a randomly-generated relevance map and two drones with their visual coverage; (\subref{fig:input:2}) the shared temporal relevance map, centered and rotated on drone 0; (\subref{fig:input:3}) the visual coverage of drone 0; (\subref{fig:input:4}) the visual coverage of all drones, as seen from drone 0. Network input for the greedy strategy is composed of (\subref{fig:input:2}) and (\subref{fig:input:3}) stacked; network input for the cooperative strategy also adds (\subref{fig:input:4}) to the stack.}
	\label{fig:input}
\end{figure*}

For the greedy strategy, the network input is $\{s, \mathcal{M}_\mathbf{s}^t\}$, this is the drone state and the current shared temporal relevance map, and the output is a $\mathbb{R}^{12}$ vector with the estimated $Q(\{s,\mathcal{M}_\mathbf{S}^t\}, a)$ values for each possible action $a$. However, rather than modeling the state $s$ as a tuple, we propose to represent it visually, with a binary image showing the visual coverage $V(s)$ of the drone. This way, the information about the observed area is immediately available from the input data and does not have to be estimated from the agent state tuple, thus achieving a simpler model and a faster convergence rate. In order to simplify the input, the two images are centered on the drone position, rotated by the drone orientation angle, and cropped to a fixed size. This way, the network input represents the surrounding of the drone, aligned with its orientation. The two images are finally stacked to form a $64\times 64\times 2$ input tensor (Figure~\ref{fig:input}).

For the cooperative strategy, we adopted a similar representation to model the network input $\{\mathbf{S}, \mathcal{M}_\mathbf{s}^t\}$. A new binary image is created, representing the visual coverage of all the drones, and it is added to the input with the same centering, rotation and cropping procedures described above, in order to create a $64\times 64\times 3$ input tensor. The choice of keeping the representations of $V(s)$ and $V(\mathbf{S})$ separatedly (second and third elements in the input stack respectively) rather than relying on  $V(\mathbf{S})$ alone (which contains $V(s)$) is used to feed the network with an explicit information to discriminate between the drone on which the network is being executed and the remaining ones. The final input for cooperative mode is shown in Figure~\ref{fig:input}.

\begin{figure}
	\centering
	\includegraphics[width=0.9\linewidth]{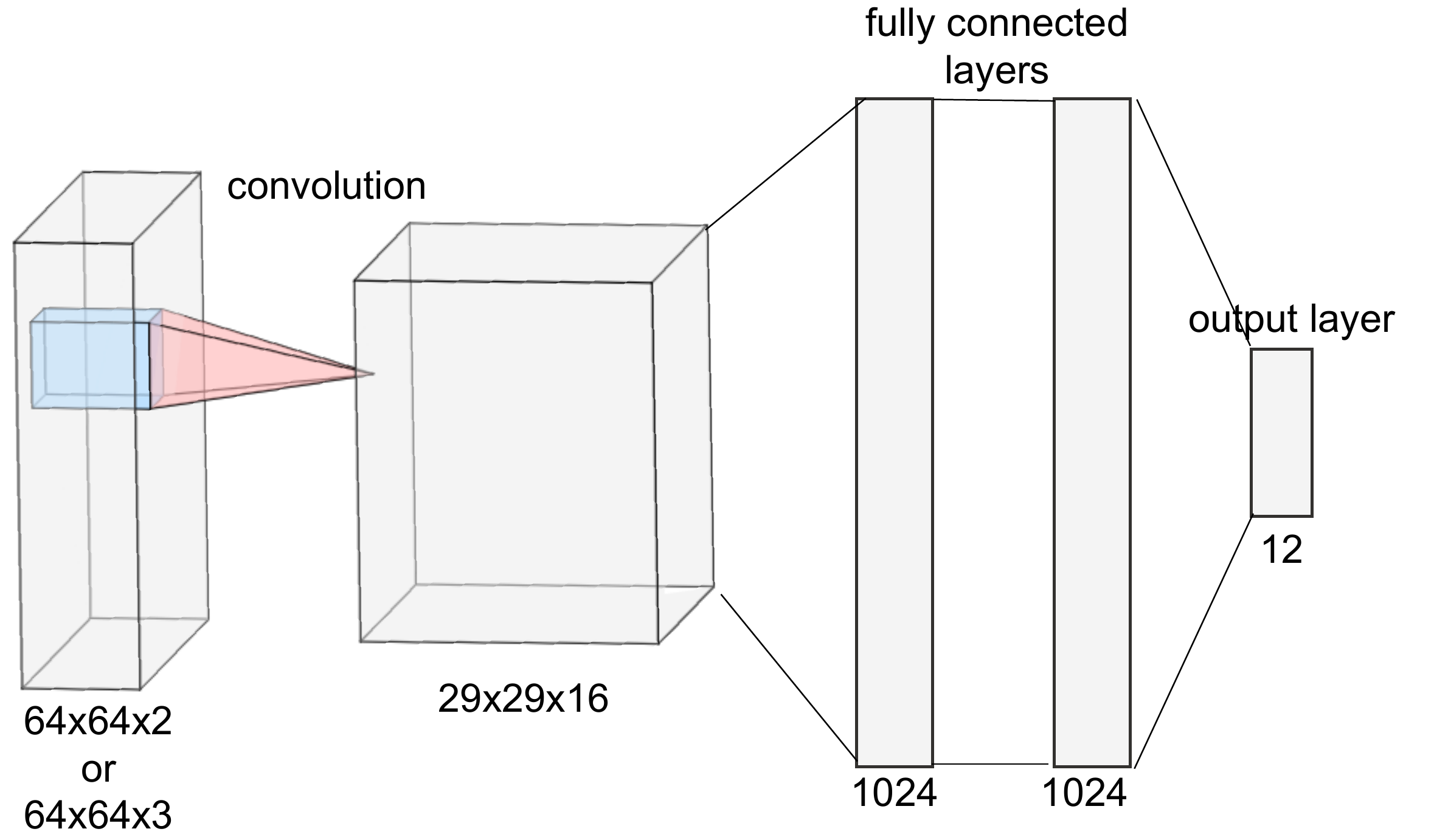}
	\caption{The proposed network structure.}
	\label{fig:network}
\end{figure}

The rest of the network is the same for the two approaches, and consists into a single convolutional layer, composed of 16 $8\times 8$ filters with stride=2. The convolutional layer is followed by two $1024\times 1$ fully connected layers and a $12\times 1$ output layer representing the $Q$ values for all the possible drone actions. All the layers use ReLU activation functions, except for the output layer, where the activation function is linear. The network topology is shown in Figure~\ref{fig:network}.

The training is done by creating a random relevance map and a random number of drones, with random initial states, at each epoch. The system then evolves for 20 steps (this is, 20 actions for each drone), either by choosing random actions or by using the partially trained network (exploration/exploitation). During this procedure, old and new states, actions and rewards are stored in the replay memory. After 20 steps, a random batch of samples is extracted from the replay memory and used to compute the loss function and train the network. The process is repeated for every epoch.

\section{Experimental results}
\label{sec:results}

The proposed model has been implemented and tested in Python using Tensorflow and the Keras library. As described in section~\ref{subsec:deepimplementation}, the adopted model is a Double Deep Q-Network with soft update of the second network and replay memory. Each epoch consists in the creation of a random relevance map and the instantiation of a random number of drones, between 1 and 5, with random valid initial states. The system evolves for 20 steps, during which each done collects the tuples $(s, s', a, r(s,a))$ and stores them in the replay memory. The actions are chosen according to the exploration/exploitation strategy, starting with 100\% exploration and linearly decreasing it until 10\% exploration is used at the end of the training. At the end of each epoch, a batch set of tuples is extracted from the replay memory with uniform probability distribution and it is used for training. The full set of training hyperparameters is shown in table~\ref{tab:hyper}. Training and all the experiments have been computed on a dual Xeon E5-2660 CPU, 224 GB RAM, 1 Tesla K40 and 2 Titan XP GPUs. 
\begin{table}[]
	\centering
	\caption{Deep Q-Network training hyperparameters.}
	\label{tab:hyper}
	\begin{tabular}{ll}
		\toprule
		Hyperparameter & Value \\ \midrule
		Discount factor $\gamma$ (eq. \ref{eq:optimalpolicy}) & 0.99 \\
		Replay memory size & $10^4$ \\
		Batch size & 64 \\
		Optimizer & Adam \\
		Adam learning rate & $10^{-4}$ \\
		Epochs (greedy) & $10^5$ \\
		Epochs (coop) & $3\times 10^5$ \\
		Soft update $\alpha$ (eq. \ref{eq:softupdate}) & 0.001 \\
	\end{tabular}
\end{table}

\subsection{Convergence to high-relevance zones}
We started the evaluation process by checking if the proposed approach can really control a drone so that it can always converge to high-relevance zones. In order to do the test, we relied on the greedy network and used it in a single-drone configuration with the temporal map update disabled. This way, the expected behavior is that the drones will converge towards the zone with highest relevance and, once reached, they will keep patrolling the same zone. 

\begin{figure}
	\centering
	\includegraphics[width=0.7\linewidth]{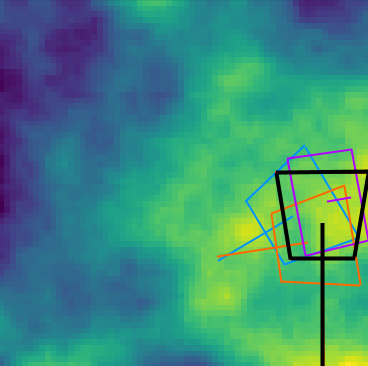}
	\caption{The optimal configuration found by brute force (black) and some approximations computed by the proposed method (other colors). For each configuration  the visual coverage (trapezoidal area) and the line connecting the drone with the projection of the image center are shown.}
	\label{fig:bf}
\end{figure}

In order to have a reference measure, as a first experiment we generated a random relevance map (Figure~\ref{fig:bf}) and searched for its optimal COR (eq.~\ref{eq:cor}) using a brute-force, exhaustive search. Since each one of the six state parameters has been discretized in 32 values, the total amount of possible states is $32^6$ and its exhaustive search would require approximately 280 hours if performed on the adopted hardware. In order to make the problem tractable, we halved the number of discretization steps of each state parameter, thus finding the brute-force optimal COR=0.079 in approximately 2.8 hours. 

We then executed the proposed method 100 times on the same map, with random initial conditions and measured the final COR to check if it was similar to the brute force one. On average, the final COR was 0.075, thus 95\% of the best possible result. The proposed method can thus find drone configurations with a coverage that is close to the best possible one, although in a fraction of time: on average, each run found the optimal COR in 0.61 seconds. Figure~\ref{fig:bf} shows the best result as well as some of the approximated solutions.
 
It is also interesting to measure the efficiency of the proposed method in finding the best solution efficiently, this is in a small number of actions. We thus measured the number of actions required by each run in order to find the final solution, and divided it by the minimum number of steps possible, which is easily computed once the initial and ending states are known. The average ratio is 1.54, thus we can expect the proposed method to reach the optimal configuration in approximately 50\% more steps than the best possible path in the state space.

\begin{figure}
	\centering
	\includegraphics[width=0.9\linewidth]{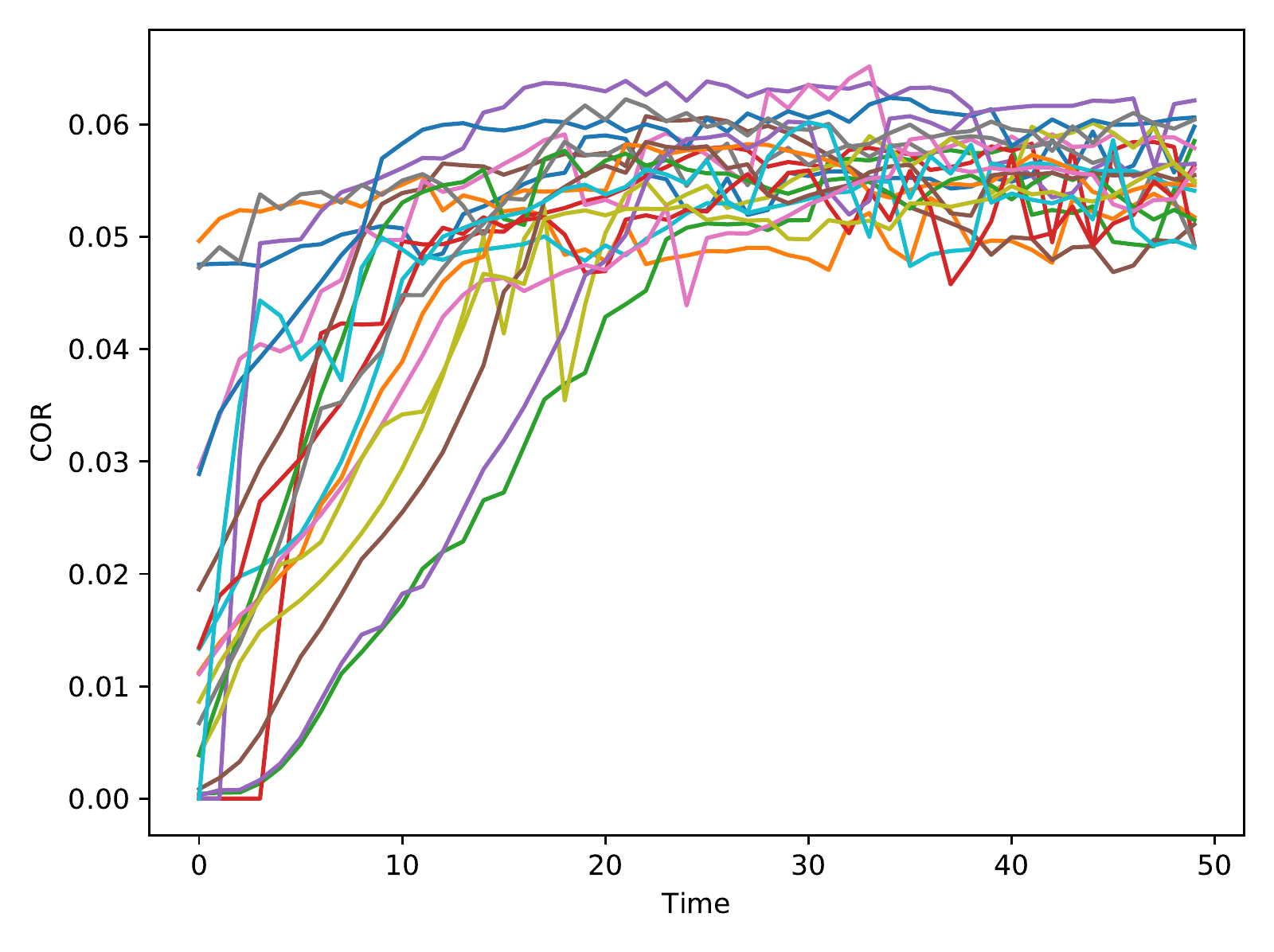}
	\caption{COR values of 20 drones running on the same map for 50 steps (no temporal update of the map).}
	\label{fig:multi_cor}
\end{figure}

The proposed experiment however used a single relevance map, and it is not suited for computing robust results, since each brute force computation is extremely time-consuming. We thus developed another test that can be more easily applied on several maps. On each map, we ran 20 agents for 50 time intervals and measured the COR of each drone. The starting state of each drone is chosen randomly. If the proposed algorithm works, we expect all the COR values to converge to a similar final score, meaning that every drone has reached an approximation of the global maximum. Figure~\ref{fig:multi_cor} shows this behavior, where it is clearly seen that after roughly 20 steps all the drones reached stable and similar COR values. Small fluctuations still exist, since our model do not explicitly considers the ``do nothing'' action (to avoid getting stuck) and thus the drone keeps moving around the found optimal area.

 In order to measure this behavior numerically, we analyzed the final COR values by normalizing them so that their mean is set to 1 and by computing their standard deviation $\sigma$. The normalization step is required to get comparable results between different maps. We repeated the test on 50 different relevance maps, and the results are shown in table~\ref{tab:multi_cor}. As it can be seen, the final standard deviation is relatively small, with an average value of $\sigma=0.13$. This means that the majority of the final COR values lie within their mean $\pm$ 13\%, thus proving the  capability of the system to reach high-relevance areas.

\begin{table}[]
	\centering
	\caption{Standard deviations of 20 normalized COR values on 40 different relevance maps.}
	\label{tab:multi_cor}{
	\begin{tabular}{ll|ll|ll|ll}
	\toprule
	test \# & $\sigma$ & test \# & $\sigma$ & test \#& $\sigma$ & test \# & $\sigma$ \\ \midrule
0 & 0.0547 & 10 & 0.1074 & 20 & 0.0749 & 30 & 0.0978 \\
1 & 0.2317 & 11 & 0.2127 & 21 & 0.1111 & 31 & 0.0721 \\
2 & 0.1001 & 12 & 0.0888 & 22 & 0.1852 & 32 & 0.1821 \\
3 & 0.1335 & 13 & 0.1210 & 23 & 0.1509 & 33 & 0.0705 \\
4 & 0.1966 & 14 & 0.1501 & 24 & 0.1745 & 34 & 0.1955 \\
5 & 0.1049 & 15 & 0.0753 & 25 & 0.1471 & 35 & 0.1264 \\
6 & 0.0909 & 16 & 0.1074 & 26 & 0.0933 & 36 & 0.1667 \\
7 & 0.1646 & 17 & 0.0975 & 27 & 0.1346 & 37 & 0.1798 \\
8 & 0.0729 & 18 & 0.1966 & 28 & 0.1887 & 38 & 0.1800 \\
9 & 0.0658 & 19 & 0.1106 & 29 & 0.2126 & 39 & 0.0795 \\ 
\end{tabular}}{}
\end{table}

\subsection{Single drone patrolling}
\begin{figure}
	\centering
	\begin{subfigure}[t]{0.5\textwidth}
		\centering
		\includegraphics[width=0.7\textwidth]{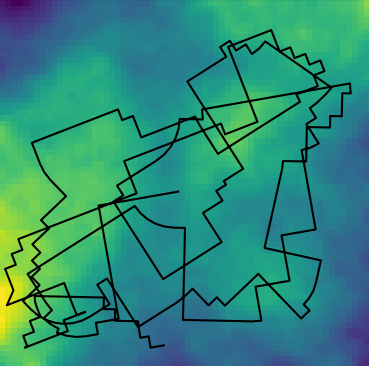}
		\caption{\label{fig:patrolling:1}}
	\end{subfigure}\\
	\begin{subfigure}[t]{0.5\textwidth}
		\centering
		\includegraphics[width=0.9\textwidth]{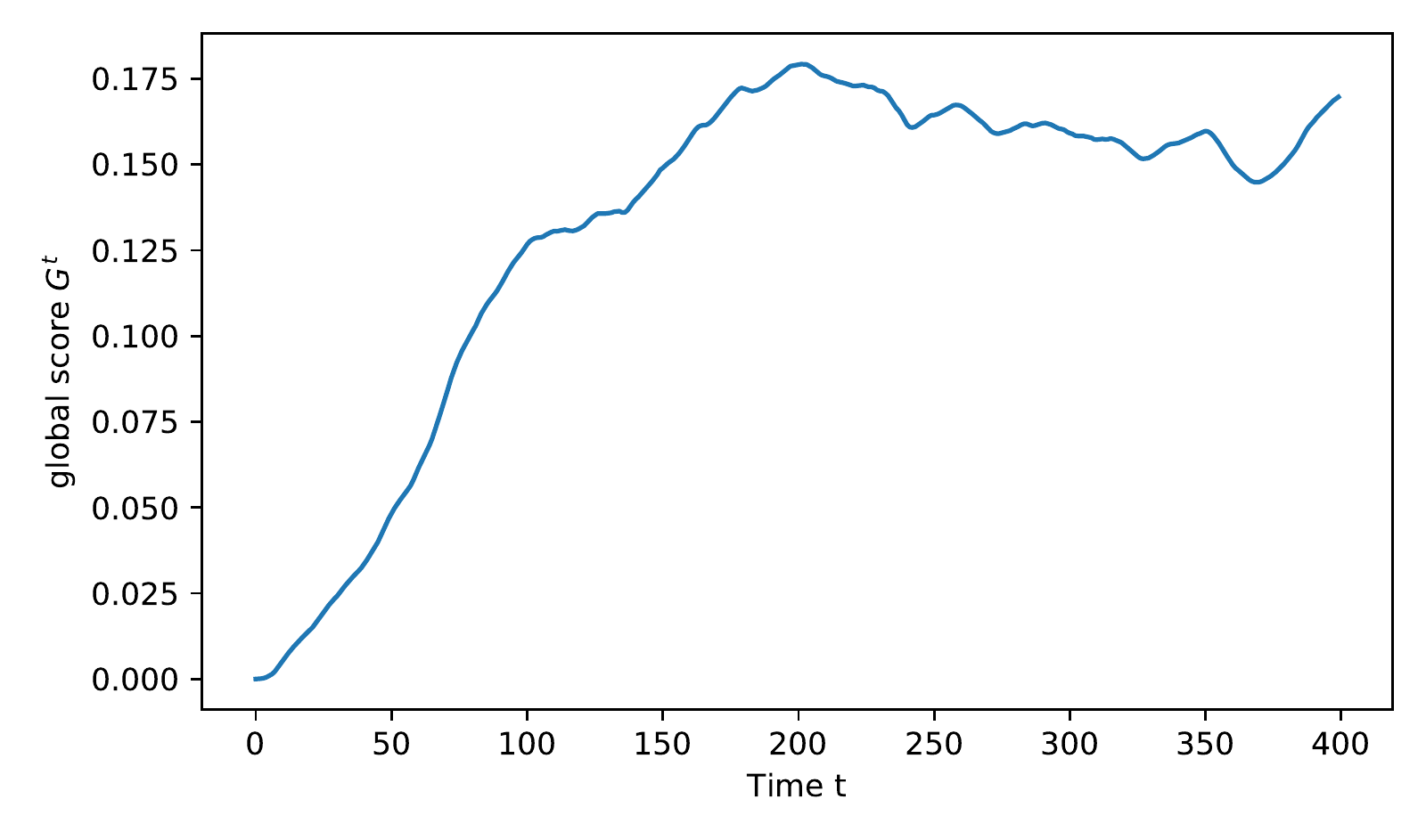}
		\caption{\label{fig:patrolling:2}}
	\end{subfigure}%
	\caption{(\subref{fig:patrolling:1}) temporal evolution of the image center of a drone, projected on the ground plane, while in patrolling mode; (\subref{fig:patrolling:2}) the corresponding global coverage $G^t$.}
	\label{fig:patrolling}
\end{figure}

As mentioned in section~\ref{subsec:patrolling}, patrolling behavior is achieved by enabling the temporal update of the relevance map. With temporal update, observed areas gradually decrease their relevance, thus making more convenient to move to a different state in search for higher rewards. With patrolling enabled, a drone should thus avoid static behaviors in which it keeps monitoring always the same zone. Figure~\ref{fig:patrolling:1} shows the path of a drone image center, projected on the ground plane, with patrolling enabled. As it can be seen, the non-static behavior is evident. 

In order to measure the overall quality of the patrolling path, we define a global measure of patrolling quality, called \emph{global coverage} $G$ with values in $[0,1]$, as:
\begin{equation}
	G^t = 1 - \sum_{(x,y)} \frac{ \mathcal{M}^t_s(x,y) }{ \mathcal{M}^0_s(x,y) }
	\label{eq:globalcoverage}.
\end{equation}
The summation part of eq.~(\ref{eq:globalcoverage}) is the ratio between the current relevance of the entire map at time $t$ and the initial total relevance (this is, the relevance map without temporal updates). The ratio thus has a maximum value of 1, and lower values denote a good patrolling, since it means that the drone covered high-relevance areas making their temporal relevance decrease. $G^t$ is then defined in order to have higher scores for good patrolling patterns. Figure~\ref{fig:patrolling:2} shows the global coverage score for the experiment shown in Figure~\ref{fig:patrolling:1}. After 200 iterations, $G^t$ reaches a stable value around 0.15, meaning that the total temporal relevance for $t>200$ is approximately 85\% of the original relevance. This measure alone is not meaningful, since it depends on many factors such as the size of the map, the speed of temporal updates, etc., however it is useful for comparative results.

\begin{figure}
	\centering
	\includegraphics[width=0.6\linewidth]{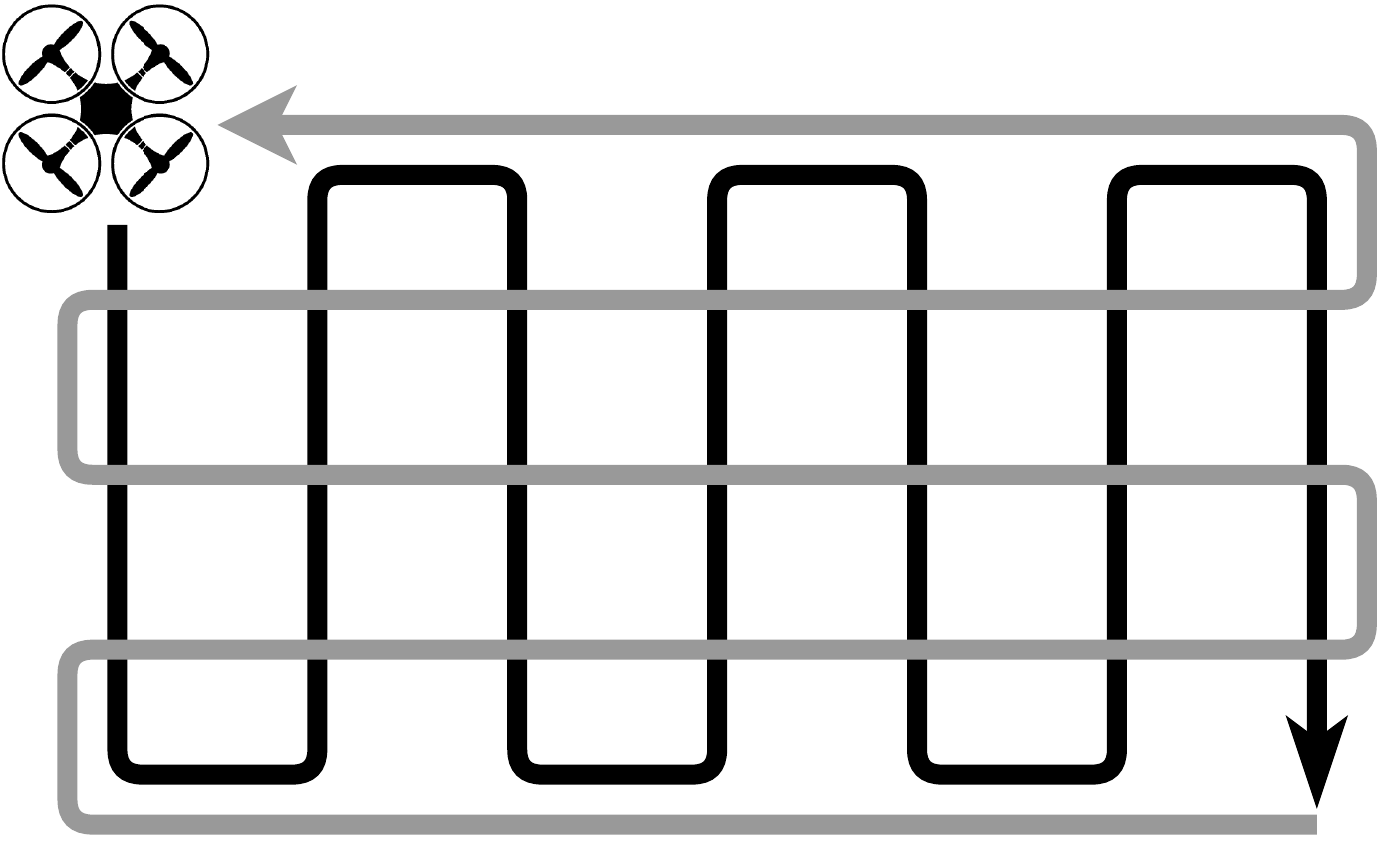}
	\caption{Naive patrolling pattern. After a sequence of vertical (black) and horizontal (grey) scans, the drone returns to its starting position and repeats the same pattern.}
	\label{fig:zigzag}
\end{figure}

\begin{figure}
	\centering
	\includegraphics[width=0.9\linewidth]{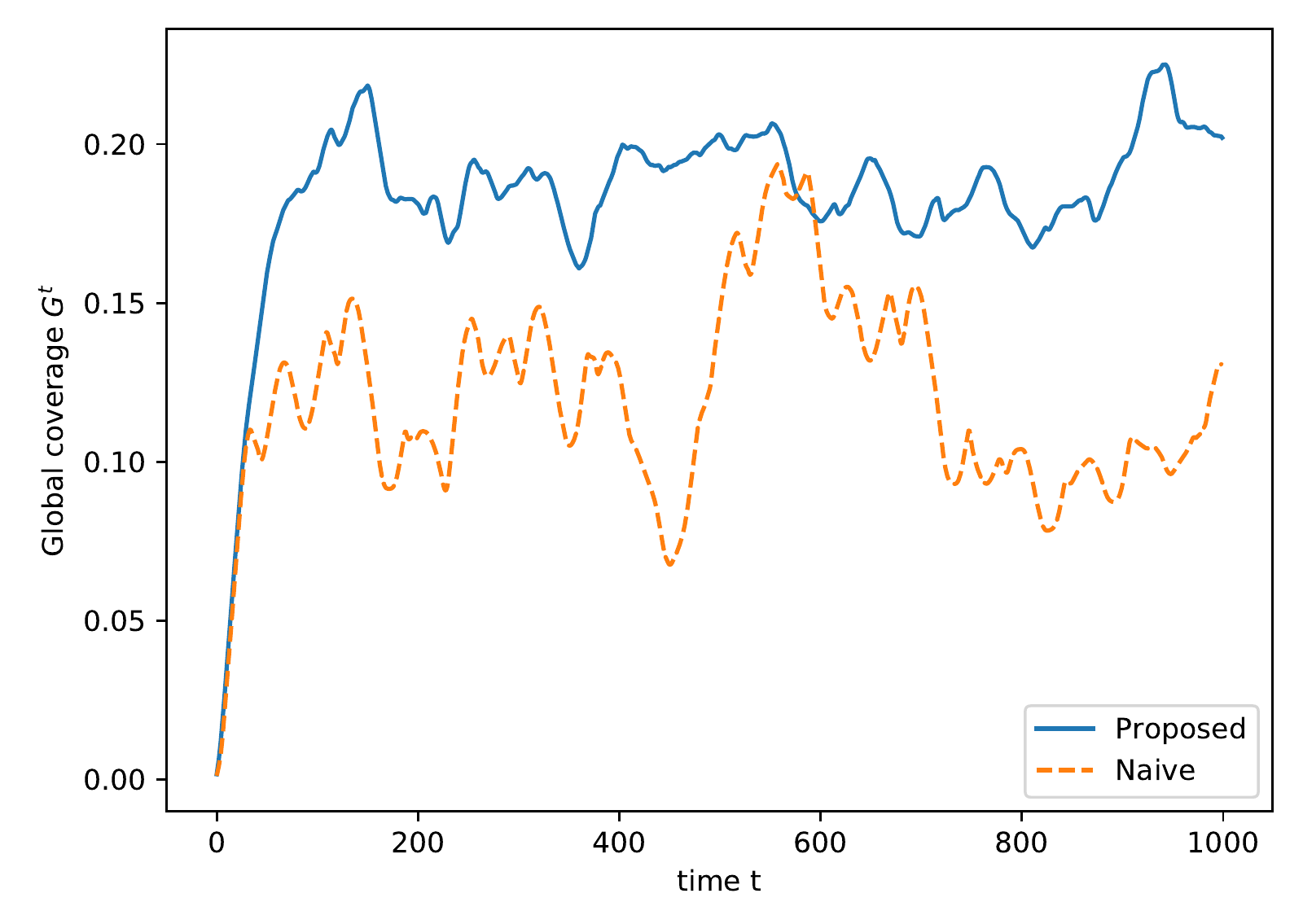}
	\caption{Global coverage scores $G^t$ for the proposed and zigzag patrolling schemes (single drone).}
	\label{fig:zigzag_test}
\end{figure}

The proposed system is thus compared with a standard, naive zigzag patrolling scheme. In this case the drone follows a predefined path, disregarding the relevance values, uniformly spanning all the environment following a path like the one shown in Figure~\ref{fig:zigzag}, alternating vertical and horizontal scans. Figure~\ref{fig:zigzag_test} shows the global score over the same relevance map of both the proposed approach and the zigzag pattern. As it can be seen, the proposed method  outperforms the naive strategy, with an average $G^t = 0.184$, more than 50\% better than the average $G^t=0.119$ obtained by the naive patrolling. Table~\ref{tab:patrolcomparison} shows the results of 50 tests on different maps, reporting the average $G^t$ for both methods and the consequent performance boost. The behavior shown in figure~\ref{fig:zigzag_test} is confirmed, and the average achieved performance boost of the proposed method is 41.95\%.

\begin{table}[]
	\centering
	\caption{Comparison of the average global coverage of the proposed method and naive zigzag patrolling schemes on 50 different maps. Each test lasts 500 steps. The boost\% column shows the improvement of the proposed vs. zigzag approach.}
		\label{tab:patrolcomparison}
		\begin{tabular}{llll|llll}
			\toprule
			test \# & ours   & zigzag & boost\% & test \# & ours   & zigzag & boost\% \\ \midrule
			1   & 0.1836 & 0.1161 &58.16    & 26  &0.1546 & 0.1192 &29.66 \\
			2   & 0.1671 &0.1162  &43.73    &27   &0.1581 & 0.1192 &32.67 \\
			3   & 0.1421 &0.1176  &20.89    &28   &0.1649 &0.1250 &31.92 \\
			4   & 0.1800 &0.1180  &52.56    &29   &0.1648 &0.1213 &35.81 \\
			5   & 0.1788 &0.1258  &42.08    &30   &0.1557 &0.1284 &21.28 \\
			6   & 0.1551 &0.1229  &26.17    &31   &0.1701 &0.1219 &39.51 \\ 
			7   & 0.1650 &0.1164  &41.71    &32   &0.1481 &0.1177 &25.77 \\
			8   & 0.2005 &0.1146  &74.97    &33   &0.1870 &0.1213 &54.24 \\
			9   & 0.1555 &0.1132  &37.39    &34   &0.1679 &0.1235 &35.99 \\ 
			10  & 0.2203 &0.1137  &93.88    &35   &0.1473 &0.1173 &25.61 \\
			11  & 0.1580 &0.1162  &35.92    &36   &0.1675 &0.1263 &32.65 \\
			12  & 0.1832 &0.1191  &53.87    &37   &0.1790 &0.1162 &54.06 \\
			13  & 0.1697 &0.1220  &39.08    &38   &0.1571 &0.1199 &31.06 \\ 
			14  & 0.1315 &0.1230  &6.86     &39   &0.1693 &0.1203 &40.68 \\
			15  & 0.1759 &0.1188  &48.09    &40   &0.1563 &0.1251 &24.97 \\
			16  & 0.1715 &0.1251  &37.03    &41   &0.1492 &0.1163 &28.35 \\
			17  & 0.1545 &0.1119  &38.04    &42   &0.1685 &0.1236 &36.36 \\
			18  & 0.1817 &0.1213  &49.80    &43   &0.1594 &0.1190 &33.91 \\
			19  & 0.1566 &0.1176  &33.13    &44   &0.1861 &0.1156 &61.01 \\
			10  & 0.2098 &0.1129  &85.80    &45   &0.1749 &0.1208 &44.79 \\
			21  & 0.1671 &0.1137  &46.98    &46   &0.1592 &0.1200 &32.69 \\
			22  & 0.1705 &0.1191  &43.15    &47   &0.1659 &0.1217 &36.36 \\
			23  & 0.1565 &0.1198  &30.59    &48   &0.1973 &0.1248 &58.05 \\
			24  & 0.1877 &0.1159  &61.92    &49   &0.1868 &0.1236 &51.14 \\
			25  & 0.1853 &0.1232  &50.42    &50   &0.1735 &0.1179 &47.19 \\	
	\end{tabular}
\end{table}

\subsection{Swarm patrolling}
In section~\ref{sec:swarm} two swarm patrolling models have been proposed, a greedy approach and a cooperative one. The greedy approach consists on each drone acting independently from the others, since the only collaboration happens indirectly by sharing the same temporal relevance map. The cooperative one instead uses an explicit representation of the states of any other drone that can be used to explain and predict temporal map changes and thus plan a better patrolling strategy. Since this predictive feature has not been explicitly coded, we rely on reinforcement learning to learn it from experience, and thus comparative results are required to show if a performance gap between the two approaches actually exists. 

\begin{figure}
	\centering
	\includegraphics[width=0.9\linewidth]{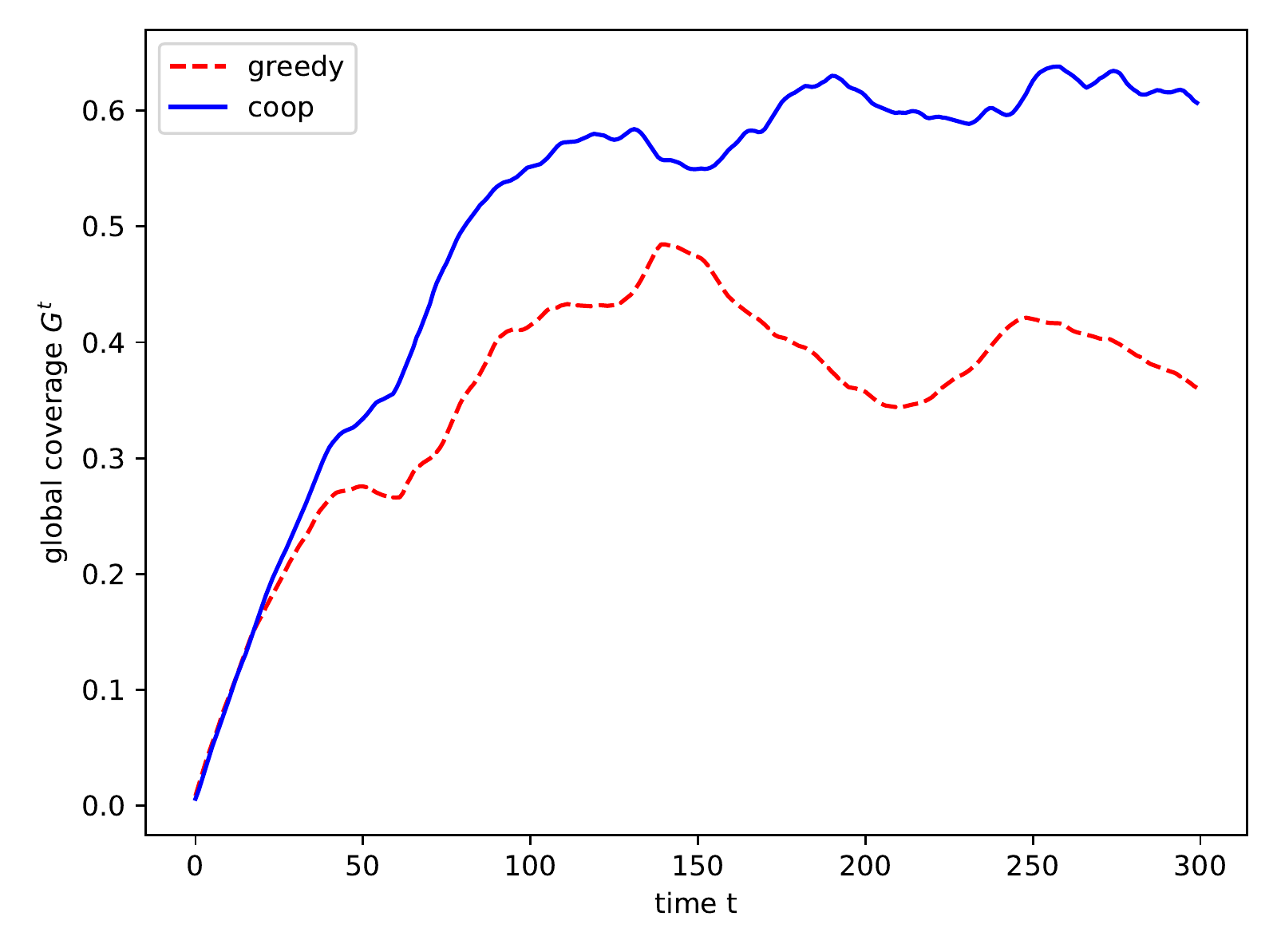}
	\caption{Global coverage scores $G^t$ of the greedy and cooperative swarm models with 4 drones, on the same map and same initial drone states.}
	\label{fig:swarm}
\end{figure}

The global coverage~(eq. \ref{eq:globalcoverage}) can be used again as a quality metric of the computed patrolling paths. Figure~\ref{fig:swarm} shows the global coverages of the greedy and cooperative approaches on the same map, with 4 drones starting from the same initial states. The cooperative approach indeed seems to perform better than the basic greedy approach, with an average global coverage of 0.505 and 0.355 respectively.

In order to confirm this result, more experiments have been conducted, as shown in table~\ref{tab:swarmcomparison}. In this case we considered a swarm of 2, 3, 4, and 5 drones respectively. For each case, we ran 20 tests on random relevance maps. In each test, both the greedy and cooperative methods were applied for 500 steps before measuring the average global coverage. The table reports the results of the two approaches, as well as the boost improvement of the cooperative approach versus the greedy one. As it can be seen, in case of multiple drones the cooperative approach always outperform the greedy one by a factor of 40-50\% on average.

In the end, the only advantage of the greedy approach is its faster training time. The cooperative approach, in order to process the additional input properly, required 3 times the number of epochs of the greedy one (see table~\ref{tab:hyper}). The greedy network however is still useful for single-drone scenarios.

\begin{table*}[tb]
	\centering
	\caption{Comparison of the average global coverage of the greedy and cooperative swarm patrolling schemes. Tests have been made with a swarm of 2, 3, 4 and 5 drones. Each test lasts 500 steps. The boost\% column show the improvement of the cooperative vs. greedy approach.}
		\footnotesize
		\label{tab:swarmcomparison}
		\begin{tabular}{l|lll|lll|lll|lll}
			\toprule
			 & \multicolumn{3}{l}{2 drones} & \multicolumn{3}{l}{3 drones} & \multicolumn{3}{l}{4 drones} & \multicolumn{3}{l}{5 drones} \\
			test \# & coop   & greedy & boost\% & coop  & greedy  & boost\% &  coop  & greedy  & boost\% & coop  & greedy  & boost\%  \\ \midrule
			1 & 0.3097 & 0.2050 & 51.06 & 0.4028 & 0.2878 & 39.94 & 0.4459 & 0.3505 & 27.22 & 0.5241 & 0.3571 & 46.75 \\
			2 &0.3111 & 0.2604 & 19.46 & 0.4877 & 0.2655 & 83.70 & 0.4940 & 0.3761 & 31.34 & 0.5767 & 0.3591 & 60.56 \\ 
			3 &0.3079 & 0.1901 & 61.95 & 0.4488 & 0.3231 & 38.91 & 0.5569 & 0.3857 & 44.40 & 0.5617 & 0.3771 & 48.96 \\
			4 &0.2714 & 0.2037 & 33.23 & 0.4524 & 0.2668 & 69.59 & 0.5137 & 0.3866 & 32.89 & 0.5363 & 0.4203 & 27.58 \\
			5 &0.2854 & 0.2159 & 32.18 & 0.4453 & 0.3417 & 30.33 & 0.5236 & 0.3581 & 46.22 & 0.5491 & 0.3413 & 60.88 \\
			6 &0.3407 & 0.1680 & 102.77 &0.3911 & 0.2126 & 83.94 & 0.5276 & 0.3481 & 51.59 & 0.5294 & 0.3944 & 34.22 \\
			7 &0.3389 & 0.2374 & 42.76 & 0.3278 & 0.2721 & 20.48 & 0.5553 & 0.3975 & 39.69 & 0.5608 & 0.3789 & 48.02 \\
			8 &0.2268 & 0.1595 & 42.16 & 0.4549 & 0.3640 & 24.97 & 0.4761 & 0.2807 & 69.63 & 0.5823 & 0.4830 & 20.56 \\
			9 &0.3039 & 0.2301 & 32.06 & 0.4484 & 0.3238 & 38.45  & 0.5392 & 0.3862 & 39.62 & 0.6006 & 0.3642 & 64.89 \\
			10 &0.3200 & 0.1828 & 75.07 &0.4072 & 0.2503 & 62.69 & 0.4843 & 0.3707 & 30.67 & 0.5946 & 0.3728 & 59.49 \\
			11 &0.2859 & 0.1879 & 52.15 & 0.3909 & 0.2389 & 63.65  & 0.5410 & 0.3705 & 46.01 & 0.5232 & 0.3900 & 34.16 \\
			12 &0.2210 & 0.1962 & 12.66 & 0.4443 & 0.3374 & 31.67 & 0.5035 & 0.3415 & 47.46 & 0.5566 & 0.3287 & 69.36 \\
			13 &0.3431 & 0.2040 & 68.13 & 0.4229 & 0.3160 & 33.82  & 0.5300 & 0.3463 & 53.02 & 0.5823 & 0.4426 & 31.56 \\
			14 &0.3128 & 0.2626 & 19.11 & 0.4398 & 0.3021 & 45.57  & 0.4804 & 0.3563 & 34.82 & 0.6036 & 0.4150 & 45.46 \\
			15 &0.3099 & 0.2562 & 20.94 & 0.4323 & 0.2670 & 61.92  & 0.5219 & 0.3475 & 50.18 & 0.5862 & 0.4667 & 25.59 \\
			16 &0.3353 & 0.2006 & 67.19 & 0.4381 & 0.2733 & 60.27 & 0.5184 & 0.3411 & 51.98 & 0.5707 & 0.3717 & 53.55 \\
			17 &0.2623 & 0.1571 & 66.93 & 0.3967 & 0.2701 & 46.85  & 0.4918 & 0.3797 & 29.51 & 0.5870 & 0.4846 & 21.13 \\
			18 &0.3047 & 0.1575 & 93.44 & 0.4562 & 0.3159 & 44.41  & 0.5076 & 0.2988 & 69.88 & 0.5806 & 0.4320 & 34.42 \\
			19 &0.2933 & 0.2193 & 33.74 & 0.4581 & 0.3183 & 43.91  & 0.5227 & 0.3910 & 33.70 & 0.5618 & 0.3552 & 58.17 \\
			20 &0.3011 & 0.1917 & 57.10 & 0.4704 & 0.2786 & 68.81 & 0.5088 & 0.3836 & 32.62 & 0.5473 & 0.3095 & 76.84 \\ \midrule
			\textbf{avg} &0.2992 &	0.2043 & 49.20 & 0.4308 & 0.2912 & 49.69 & 0.5121 & 0.35982 & 43.12 & 0.5657 & 0.3922 & 46.10 \\  
	\end{tabular}
\end{table*}

\subsection{Computational load}
Since the proposed theoretical framework is currently implemented as a Python simulation, we cannot measure the timings of a full sense-act cycle. However we can measure the computational time required to choose an action given a state input, in order to understand if a practical implementation is actually feasible. Choosing an action requires to run the input through the neural network. This operation in our experiments consistently requires 1.5 ms on a CPU-only implementation running on an Intel(R) Core(TM) i7-9700K 3.60 GHz processor. We believe this time is negligible if compared to the time required to actually execute the chosen action.

\section{Conclusions}
In this work we proposed a theoretical model to compute efficient patrolling paths for single drones and swarms of drones. The problem of uneven coverage requirements was explicitly considered: the environment zones are associated to different priority (relevance) scores, expressing the importance for an area to be visually covered by a drone. The proposed implementation uses reinforcement learning to train a deep network that selects the best action that will most likely lead to a good coverage in the long run. The proposed system was extensively tested and showed good performances, also compared to a standard patrolling scheme.

The proposed work can be a reference framework for real-world implementations, although in that case several additional constraints should be considered, such as different topological altitudes of the environment, presence of obstacles, battery life, etc.. 

\section*{Acknowledgments}
This work is partially supported by the PNRM project ``Proactive Counter-UAV'' (a2018.045).

\bibliographystyle{iet}
\bibliography{biblio,myworks}

\end{document}